\documentclass{article}

\usepackage{microtype}
\usepackage{graphicx}
\usepackage{booktabs} % for professional tables

\usepackage{hyperref}

\newcommand{\prompt}[2][0.9]{%
  {\ttfamily
   \spaceskip=#1\fontdimen2\font
             plus #1\fontdimen3\font
             minus #1\fontdimen4\font
   \xspaceskip=\spaceskip
   #2%
  }%
}

\usepackage[preprint]{icml2026}

\usepackage{amsmath}
\usepackage{amssymb}
\usepackage{mathtools}
\usepackage{amsthm}
\usepackage{bm}
\usepackage{graphicx}
\usepackage{appendix,etoc}
\usepackage{subfigure}
\usepackage{multirow}

\def \x {\bm{x}}
\def \y {\bm{y}}
\def \p {\bm{p}}

\usepackage[capitalize,noabbrev]{cleveref}

\theoremstyle{plain}

\theoremstyle{definition}

\theoremstyle{remark}

\usepackage[textsize=tiny]{todonotes}

\icmltitlerunning{Submission and Formatting Instructions for ICML 2026}

\begin{document}

\twocolumn[
  \icmltitle{Are Multimodal Large Language Models Good Annotators for Image Tagging?}

  % It is OKAY to include author information, even for blind submissions: the
  % style file will automatically remove it for you unless you've provided
  % the [accepted] option to the icml2026 package.

  % List of affiliations: The first argument should be a (short) identifier you
  % will use later to specify author affiliations Academic affiliations
  % should list Department, University, City, Region, Country Industry
  % affiliations should list Company, City, Region, Country

  % You can specify symbols, otherwise they are numbered in order. Ideally, you
  % should not use this facility. Affiliations will be numbered in order of
  % appearance and this is the preferred way.
  % \icmlsetsymbol{equal}{*}

  \begin{icmlauthorlist}
    \icmlauthor{Ming-Kun Xie}{riken}
    \icmlauthor{Jia-Hao Xiao}{seu}
    \icmlauthor{Zhiqiang Kou}{seu}
    \icmlauthor{Zhongnian Li}{c}
    \icmlauthor{Gang Niu}{riken}
    \icmlauthor{Masashi Sugiyama}{riken,ut}
  \end{icmlauthorlist}

  \icmlaffiliation{riken}{RIKEN Center for Advanced Intelligence Project, Japan}
  \icmlaffiliation{ut}{The University of Tokyo, Japan}
  \icmlaffiliation{seu}{Southeast University, China}
  \icmlaffiliation{c}{China University of Mining and Technology, China}

  \icmlcorrespondingauthor{Ming-Kun Xie}{ming-kun.xie@riken.jp}
  % \icmlcorrespondingauthor{Firstname2 Lastname2}{first2.last2@www.uk}

  % You may provide any keywords that you find helpful for describing your
  % paper; these are used to populate the "keywords" metadata in the PDF but
  % will not be shown in the document
  \icmlkeywords{Machine Learning, ICML}

  \vskip 0.3in
]

% this must go after the closing bracket ] following \twocolumn[ ...

% This command actually creates the footnote in the first column listing the
% affiliations and the copyright notice. The command takes one argument, which
% is text to display at the start of the footnote. The \icmlEqualContribution
% command is standard text for equal contribution. Remove it (just {}) if you
% do not need this facility.

% Use ONE of the following lines. DO NOT remove the command.
% If you have no special notice, KEEP empty braces:
\printAffiliationsAndNotice{}  % no special notice (required even if empty)
% Or, if applicable, use the standard equal contribution text:
% \printAffiliationsAndNotice{\icmlEqualContribution}

\begin{abstract}
  Image tagging, a fundamental vision task, traditionally relies on human-annotated datasets to train multi-label classifiers, which incurs significant labor and costs. While Multimodal Large Language Models (MLLMs) offer promising potential to automate annotation, their capability to replace human annotators remains underexplored. 
  This paper aims to analyze the gap between MLLM-generated and human annotations and to propose an effective solution that enables MLLM-based annotation to replace manual labeling.
  Our analysis of MLLM annotations reveals that, under a conservative estimate, MLLMs can reduce annotation cost to as low as one-thousandth of the human cost, mainly accounting for GPU usage, which is nearly negligible compared to manual efforts. 
  Their annotation quality reaches about 50\% to 80\% of human performance, while achieving over 90\% performance on downstream training tasks.
  Motivated by these findings, we propose TagLLM, a novel framework for image tagging, which aims to narrow the gap between MLLM-generated and human annotations. 
  TagLLM comprises two components: Candidates generation, which employs structured group-wise prompting to efficiently produce a compact candidate set that covers as many true labels as possible while reducing subsequent annotation workload; and label disambiguation, which interactively calibrates the semantic concept of categories in the prompts and effectively refines the candidate labels.
  Extensive experiments show that TagLLM substantially narrows the gap between MLLM-generated and human annotations, especially in downstream training performance, where it closes about 60\% to 80\% of the difference.
\end{abstract}

\section{Introduction}

Image tagging \citep{chen2019multi,zhang2024recognize} is a fundamental and practical vision task that aims to annotate an image with all its relevant labels. Typical methods follow the multi-label learning paradigm \citep{liu2021emerging}, which involves training a multi-label classifier on a human-annotated dataset and then using it to predict all relevant labels for unseen images. While this framework has achieved remarkable success, it suffers from a potential limitation: The requirement of manually annotating a new dataset for each new task. This annotation process can be particularly labor-intensive and costly, especially in scenarios with a large label space with hundreds or even thousands of categories. Therefore, it is crucial to reduce the cost of manual annotation for achieving efficient image classification.

Multimodal Large Language Models (MLLMs) \citep{yin2024survey} have emerged as a transformative technology with the potential to reshape the traditional supervised learning paradigm. Trained on massive cleaned web data and synthesized data, these advanced models exhibit powerful multimodal understanding capabilities across diverse domains, and in some cases, their performance can even rival that of humans. A key question is \textit{whether MLLMs are good annotators for image tagging}. Can they produce human-level annotations and, under appropriate conditions, replace or at least complement human annotators?

% One important question arises: Are LVLMs good annotators for image tagging? Can we leverage LVLMs to replace human annotators for image tagging, with the goal of producing annotations comparable to human-level quality?

\begin{figure*}[!t]
    \centering
    \includegraphics[width=0.9\linewidth]{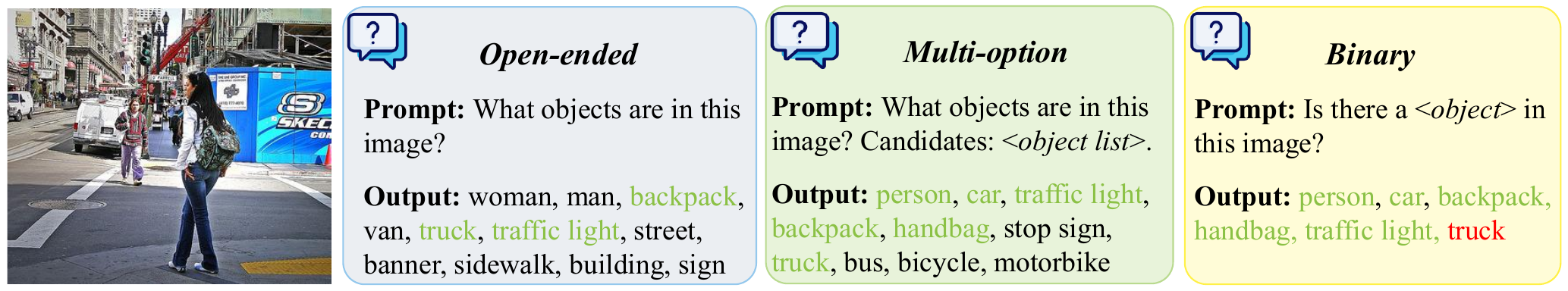}
    \caption{An example of MLLM-generated annotations using different prompting methods. {\color{green!50!black} Green} denotes true labels. {\color{red} Red} denotes missing labels. \textit{Open-ended} and \textit{multi-option} prompting require only a single inference per image, whereas \textit{binary} prompting requires $q$ inferences per image, where $q$ is the number of classes.}
    \label{fig:example}
    \vspace{-1em}
\end{figure*}

To answer this question, we conduct a systematic study to investigate the ability of MLLMs on image tagging. On the one hand, we find that prompts are crucial for annotation along two dimensions: prompt format and prompt style. The former determines the annotation mode, which mainly includes the multi-option and binary formats. The latter concerns the degree of prompt refinement, and in general, carefully-designed prompts yield higher annotation quality.
On the other hand, we evaluate the annotation ability of MLLMs from two perspectives: the quality of MLLM-generated annotations and the performance of models trained on these annotations. 
For annotation quality, we find that MLLMs achieve promising performance but still exhibit a noticeable gap compared to human annotators, reaching only about 50\% of human-level quality in the worst case.
The reason is intuitive: they perform well on common categories, whereas their performance significantly degrades on uncommon or ambiguous ones. 
Regarding downstream model performance, models trained on MLLM-generated annotations typically reach around 90\% of the performance of models trained on human annotations, and, surprisingly, even surpass them on some categories.
These findings indicate that MLLM-generated annotations offer the dual benefit of substantially reducing manual annotation costs while potentially enhancing annotation quality.

In this paper, we introduce an MLLM-driven annotation framework, TagLLM, which aims to produce human-level annotations for image tagging. Our main contribution is a two-stage pipeline: a \textit{multi-option} prompting stage that efficiently produces a compact candidate label set, typically comprising fewer than one tenth of the categories, followed by a \textit{binary} prompting verification stage that carefully refines the candidate labels and significantly enhances annotation quality. Specifically, in the first stage, we adopt a divide-and-conquer prompting strategy that groups frequently co-occurring categories together to encourage within-group competition. This method allows only high-confidence labels to stand out, thereby improving annotation precision. 
Empirically, we find that many false positives in candidate sets are caused by what we refer to as \textit{concept misalignment}, where the category name does not accurately correspond to the actual object concept. In the second stage, we introduce a concept-aligned disambiguation method, which leverages ChatGPT-4o to interactively calibrate category names for refining candidate label sets. By integrating the complementary strengths of two prompting methods, TagLLM delivers efficient operation and high-quality annotations. 
% Importantly, every stage substantially shrinks the candidate label sets, enabling us to introduce human-assisted calibration at low annotation cost and achieve human-LLM collaborative annotation, which in turn significantly enhances annotation quality.
Extensive experimental results on multiple benchmark datasets validate the effectiveness of the proposed method.

% Inspired by the idea of crowdsourcing annotation, we develop an LVLM-ensemble candidates generation strategy, which utilizes multiple LVLMs to generate a candidate label set for each image. The strategy aims to strike a balance between maximizing the true positive coverage and accepting a moderate level of label noise (false positives). Moreover, we find that many false positives are caused by what we refer to as \textit{concept disalignment}, where the category name does not accurately correspond to the actual object. To address this problem, we propose a concept alignment method, which leverages ChatGPT-4o to interactively calibrate category names for disambiguating the candidate label sets. 

\section{A Close Look to MLLM Annotations}

\label{sec2}

In this section, we present a systematic investigation of MLLM annotation for the image tagging task. We begin by introducing notation and settings.

% with the goal of characterizing how it differs from human-provided labels. We begin by introducing notation and settings.

\subsection{Notation and Settings}

We are given a dataset $\{\x_i\}_{i=1}^n$ consisting of $n$ images, where $\x_i\in\mathcal{X}$ and $\mathcal{X}=\mathbb{R}^d$ is the feature space, along with a category vocabulary $\{C_1, C_2, \ldots, C_q\}$ of $q$ category names to be annotated. Each image is associated with an unknown label vector $\y_i\subseteq\mathcal{Y}$, where $\mathcal{Y}=\{0, 1\}^q$ is the label space with $q$ possible labels. Here, $y_k=1$ indicates that the $k$-th label with the name $C_k$ is relevant while $y_k=0$ indicates that it is not. We use $[q]$ to denote the integer set $\{1, 2, \ldots, q\}$.

To disclose the characteristics of MLLM-generated annotations, we employ one of the most powerful open-source MLLMs, Qwen3-VL, to annotate two widely used multi-label image benchmark datasets, MS-COCO 2014 (COCO 2014) \citep{lin2014microsoft} and Objects365 (O365) \citep{shao2019objects365}. Given the flexibility of natural language, the number of possible prompt templates is virtually unlimited. In our study, we characterize prompts using two dimensions: \textit{prompt format} and \textit{prompt style}. We will provide a detailed discussion in the following subsection.

\begin{figure*}[!t]
\begin{minipage}[t]{0.49\textwidth}
    \centering
    \subfigure{
    \centering
    \includegraphics[width=0.6\linewidth]{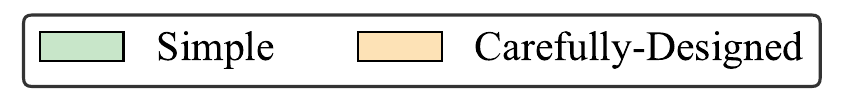}
    }
    \vspace{-0.2cm}
  
    \setcounter{subfigure}{0}
    \subfigure[BP]{
    \includegraphics[width=0.45\linewidth]{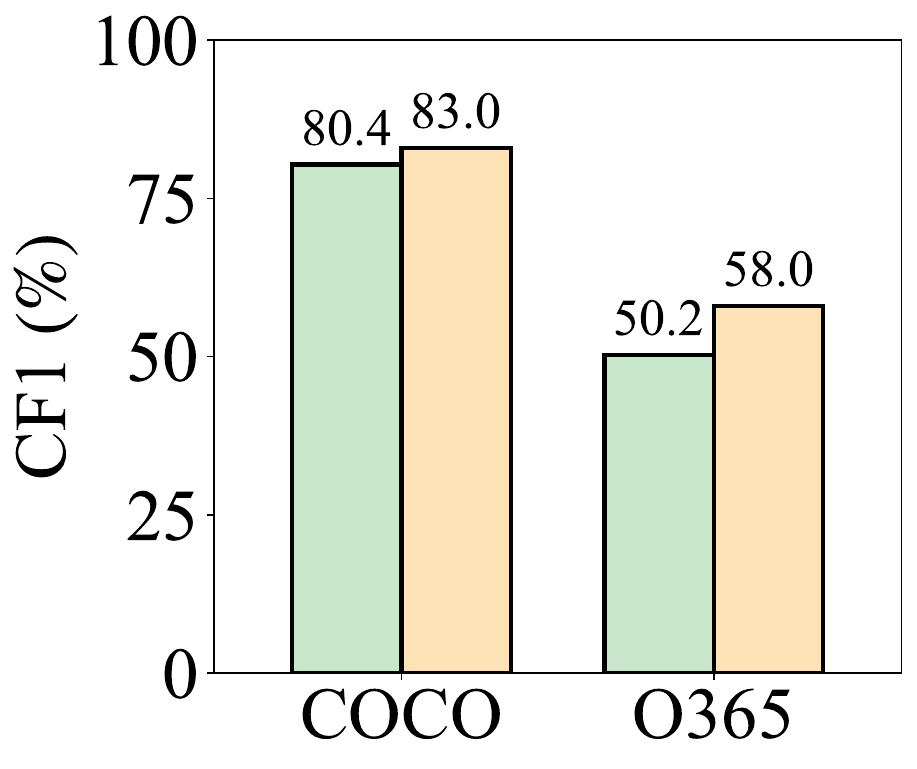}
    \label{fig:bar_cf1_bp}
    }\hfill
    \subfigure[MOP]{
    \includegraphics[width=0.45\linewidth]{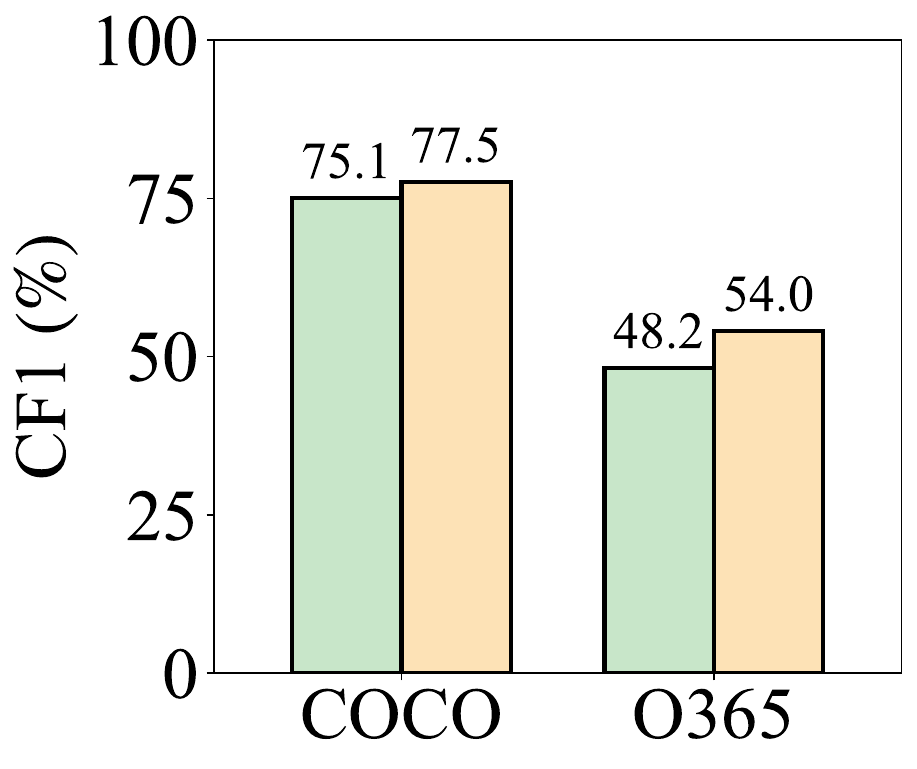}
    \label{fig:bar_cf1_mop}
    }
\end{minipage}
\begin{minipage}[t]{0.49\textwidth}
    \centering
    \subfigure{
    \centering
    \includegraphics[width=0.8\linewidth]{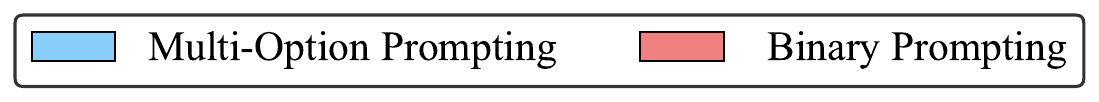}
    }
    \vspace{-0.2cm}

    \setcounter{subfigure}{2}
    \subfigure[CP]{
    \includegraphics[width=0.45\linewidth]{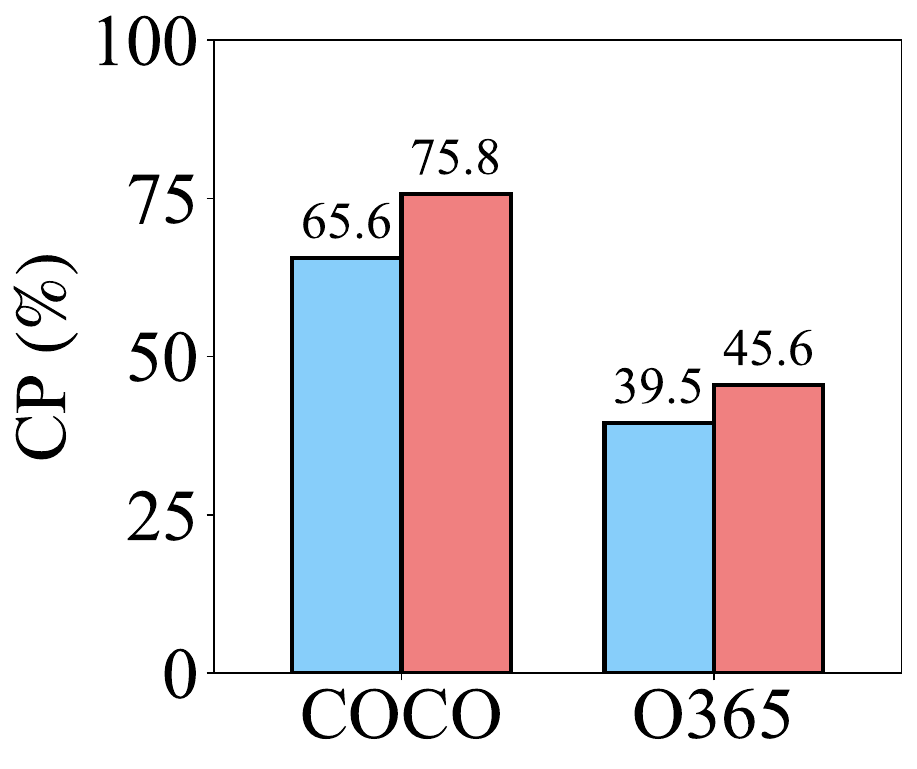}
    \label{fig:bar_cp}
    }
    \subfigure[CR]{
    \includegraphics[width=0.45\linewidth]{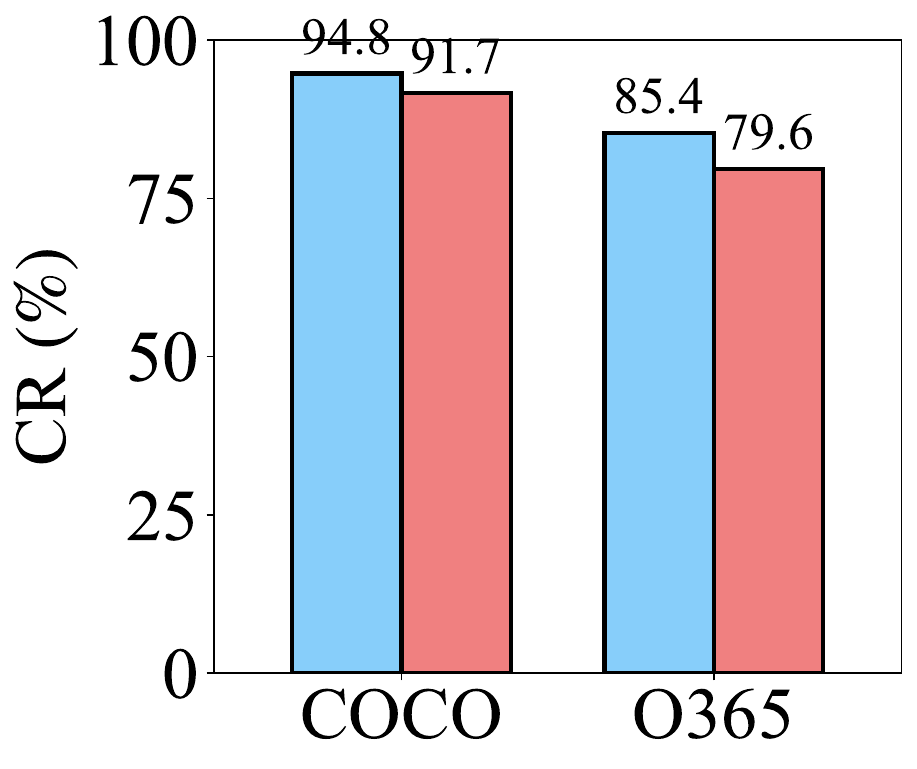}
    \label{fig:bar_cr}
    }
\end{minipage}
\caption{The quality of annotations generated by Qwen3-VL-8B using different prompt formats and prompt styles.}
\vspace{-0.5em}
\end{figure*}

\subsection{Prompt Matters}

\label{sec:prompt}

\textbf{Prompt format} specifies the task framing and defines the model’s response format, \textit{i.e.}, it constrains how the model should answer. We consider three types of prompt formats: \textit{open-ended}, \textit{multi-option} and \textit{binary}. Open-ended prompting asks MLLMs what objects are present without imposing constraints, \textit{e.g.}, \texttt{What objects are in this image?}. Multi-Option Prompting (MOP) poses the same question but restricts the answer to a predefined candidate set, \textit{e.g.}, \texttt{What objects are in this image? Candidates:<object list>.}. Binary Prompting (BP) differs from both by querying the presence of a specific object, \textit{e.g.}, \texttt{Is there a <object> in this image?}, to which the MLLM needs only answer ``Yes" or ``No". Figure \ref{fig:example} illustrates an example of the three prompt formats using Qwen3-VL-8B-Instruct. On the one hand, open-ended prompting is suboptimal for image tagging because the model often produce labels outside the candidate set, \textit{e.g.}, \textit{van}, \textit{banner}. While these labels are not strictly incorrect, they are useless for downstream model training. On the other hand, MOP and BP are better suited to the task but exhibit different behaviors: \textit{MOP tends to output more labels, covering most ground-truth ones yet introducing noise, whereas BP prioritizes precision and may miss some true labels.} This difference mainly stems from how the MLLM handles these two formats of prompts. For MOP, the model generates labels sequentially and later tokens depend on earlier ones. When visual evidence is weak (for example, due to occlusion or very small objects), the model tends to rely on textual co-occurrence and produces additional labels, \textit{e.g.}, \textit{bus} in the example. For BP, the model makes an independent Yes/No decision for a single class without conditioning on other information, so it predicts only when visual evidence is sufficient and often misses weak positives, \textit{e.g.}, \textit{truck} in the example. 

\textbf{Prompt style} describes how the prompt is phrased, including its tone, constraint strength, and level of detail, which collectively steer the model toward the desired response. As mentioned above, the simplest prompt takes the form: \texttt{Is there a <object> in this image?}. However, such a prompt is unlikely to achieve optimal annotation quality. In contrast, a more carefully-designed prompt, \textit{e.g.}, \texttt{Carefully examine the image and decide if it contains a <object>.}, can significantly improve annotation quality. For simplicity, we consider two prompt styles, \textit{simple} and \textit{carefully-designed}. For different prompt formats and prompt styles, the prompt specifications are provided in Appendix~\ref{app:prompt}.

To further investigate how prompts affect annotation quality along these two dimensions, Figure~\ref{fig:bar_cf1_bp} and~\ref{fig:bar_cf1_mop} show the annotation performance in terms of per-Class F1 (CF1) obtained using \textit{simple} and \textit{carefully-designed} prompts under different prompt formats (BP and MOP). It can be observed that, \textit{carefully-designed} prompts substantially improve annotation quality in all cases. Based on these results, we use carefully-designed prompts in the remainder of the paper unless stated otherwise. Figures~\ref{fig:bar_cp} and~\ref{fig:bar_cr} illustrate annotation results in terms of per-Class Precision (CP) and per-Class Recall (CR) using BP and MOP. These two methods show distinct annotation behaviors. MOP typically yields higher recall but lower precision, whereas BP exhibits the opposite trend, achieving higher precision but lower recall. These experimental results provide guidance for our subsequent method design.

\subsection{Comparison with Human Annotations}

\begin{figure*}[!t]
    \subfigure[Top/bottom-10 F1 scores on O365 with binary prompting.]{
        \includegraphics[width=0.5\textwidth]{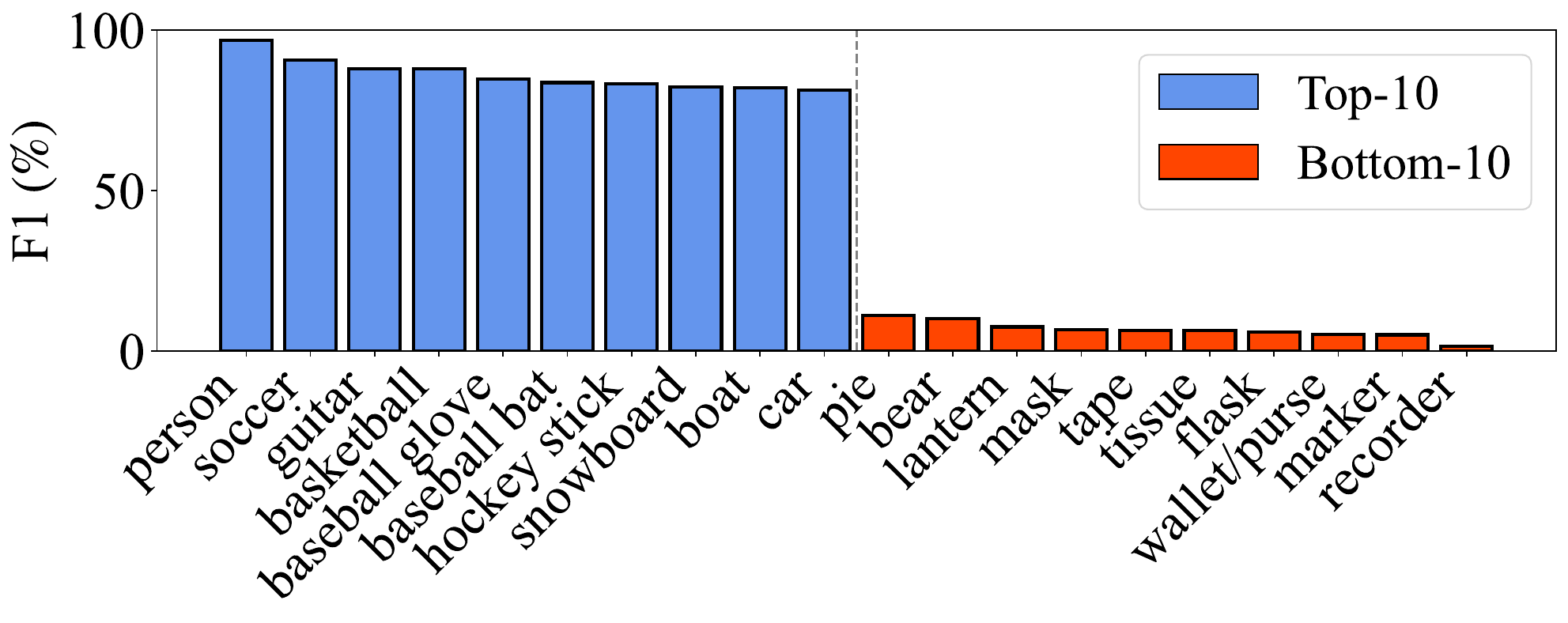}
        \label{fig:top_10}
    }
    \subfigure[Top-10 categories on O365.]{
        \includegraphics[width=0.45\textwidth]{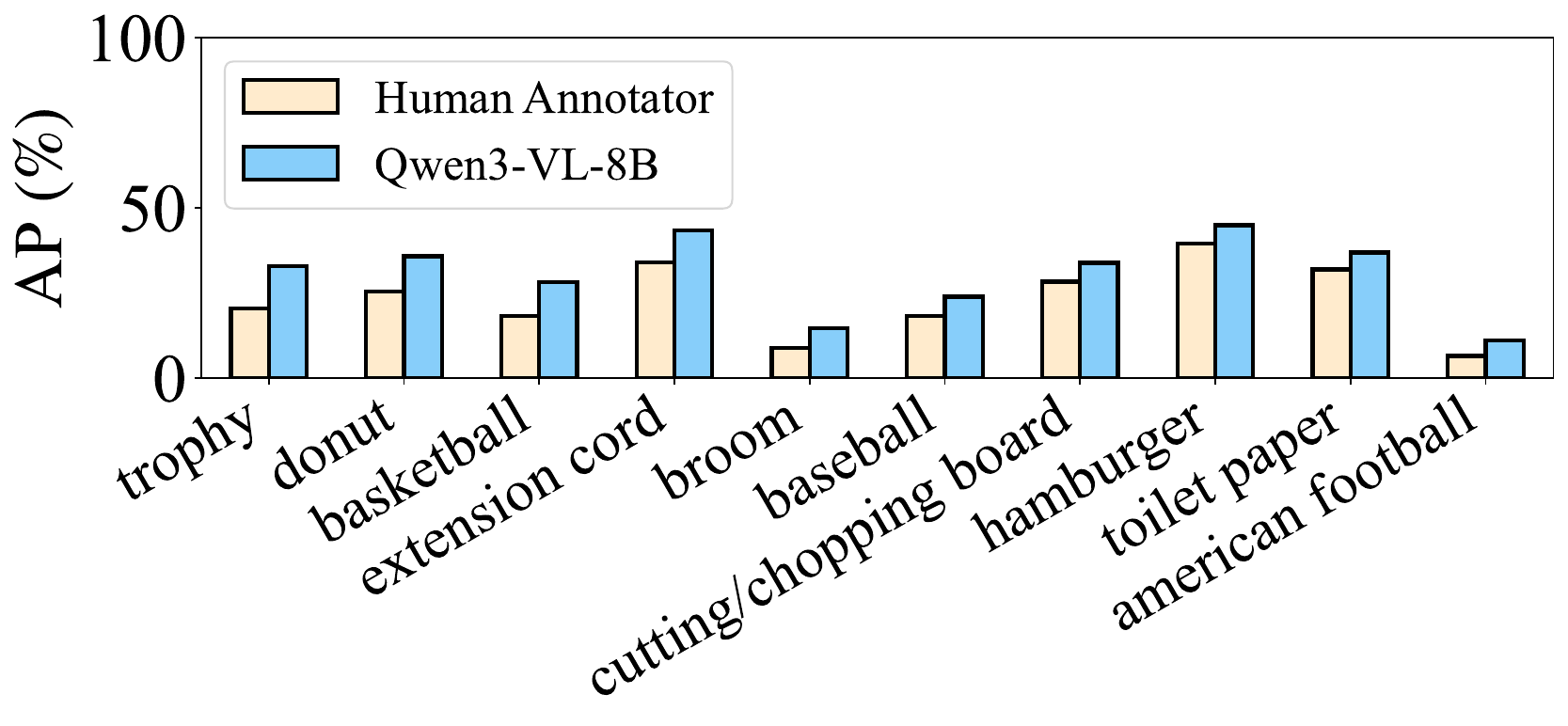}
        \label{fig:top_10_ap}
    }
    \caption{The quality of annotations and performance of training model obtained by using Qwen3-VL-8B.}
    \label{fig:annotation_quality}
\end{figure*}

Figure \ref{fig:annotation_quality} illustrates the quality of annotations generated by Qwen3-VL and the performance of models trained on these MLLM-generated annotations. Given that BP achieves higher overall annotation quality than MOP, we use BP for annotation in the these experiments. Based on these results, we summarize two core findings.

\textit{On the one hand, MLLM-generated annotations exhibit a significant disparity between common and uncommon/ambiguous categories.} From Figure~\ref{fig:bar_cp} and~\ref{fig:bar_cr}, Qwen3-VL produces high-quality annotations on COCO 2014, which primarily consists of common categories \footnote{The 80 categories in COCO 2014 were selected by the co-authors through voting based on how commonly the categories appear \citep{lin2014microsoft}.}. In contrast, on O365, which includes a large number of uncommon or ambiguous categories, the quality of the generated labels drops significantly. To provide a further validation for this phenomenon, Figure \ref{fig:top_10} illustrates the annotation quality for the top-10 and bottom-10 categories in terms of F1 scores. It can be observed that the top-10 categories are mostly common objects, such as \textit{person} and \textit{car}, or conceptually unambiguous categories like \textit{guitar}, \textit{basketball}, \textit{baseball bat}; while the bottom-10 categories tend to be ambiguous categories, \textit{e.g.}, \textit{mask}, \textit{tape}, and \textit{marker}. 
% The quality of annotations has a direct impact on model performance. 
% As shown in Figure \ref{fig:noisy_clean}, when trained on LVLM-annotated data, the model achieves 80.7\% mAP on COCO 2014 and 44.0\% mAP on O365. Compared to training on human-annotated data, the performance gap is 2.6\% on COCO and 4.6\% on O365. Notably, the performance gap on O365 is nearly twice as large as that on COCO 2014, indicating greater challenges in handling uncommon/ambiguous categories when using LVLMs for annotation. 
Overall, the current level of annotation quality cannot yet be considered comparable to human performance, and there remains significant room for improvement.

\begin{figure*}[!t]
    % \subfigure{
    %     \includegraphics[width=0.48\textwidth]{figs/bar_legend_3.pdf}
    % }

    \subfigure[]{
        \includegraphics[width=0.33\textwidth]{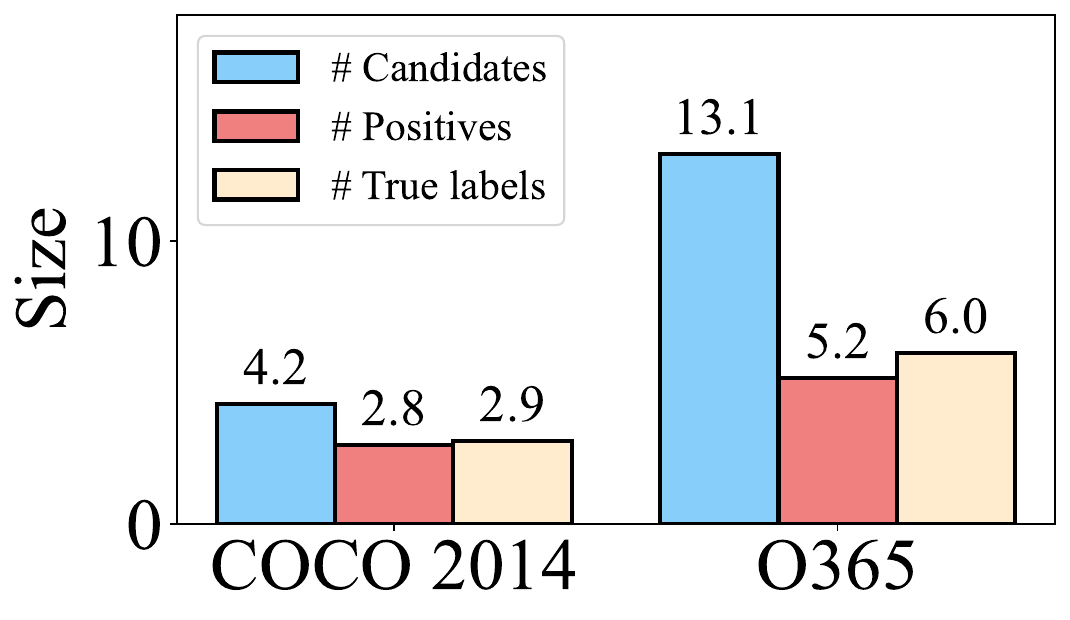}
        \label{fig:mop}
    }
    \subfigure[]{
        \includegraphics[width=0.28\textwidth]{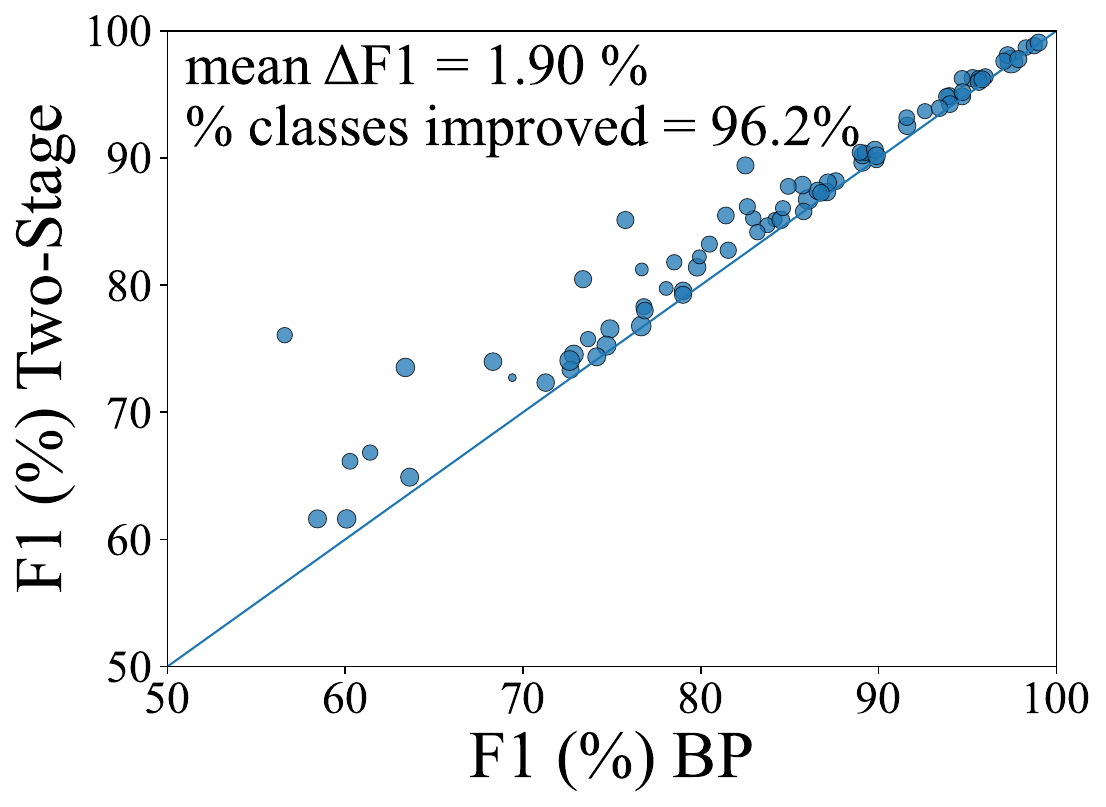}
        \label{fig:coco_f1}
    }
    \subfigure[]{
        \includegraphics[width=0.28\textwidth]{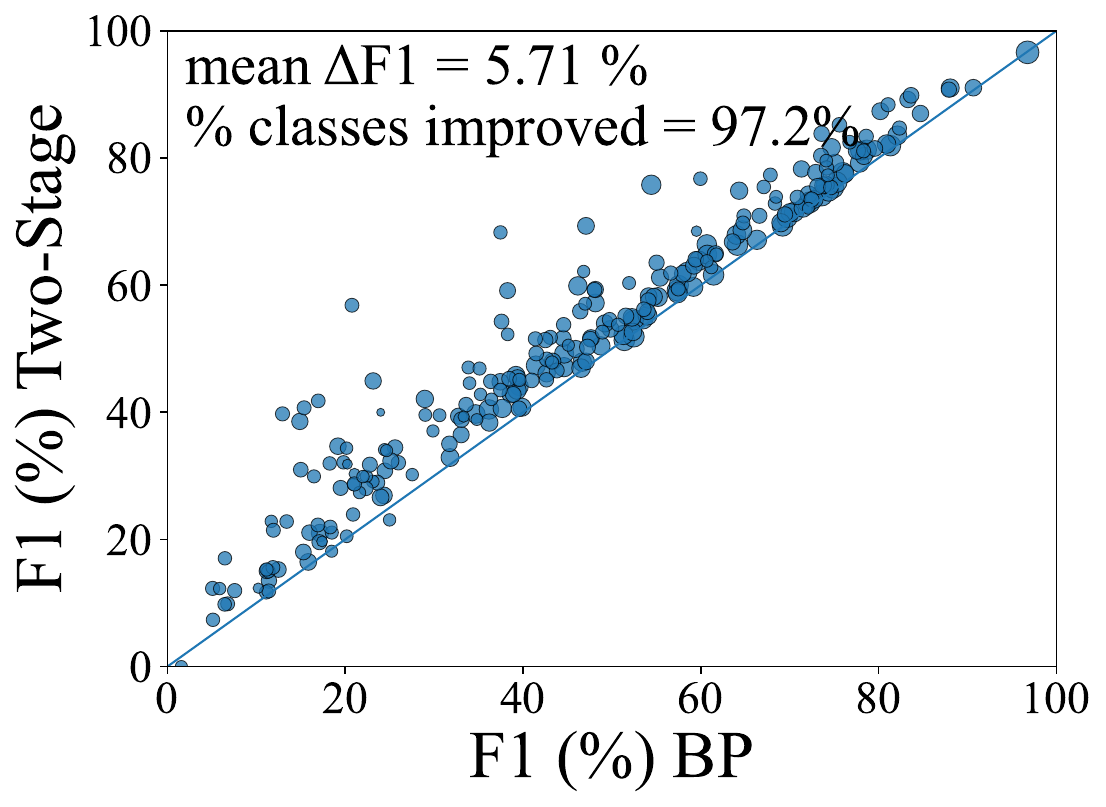}
        \label{fig:o365_f1}
    }
    \caption{Comparison of three prompting strategies on COCO and Objects365. Left: Precision; middle: Recall; right: average number of candidate labels per image. DP (Disco-occurrence Partition) prompting yields the highest recall.}
    % \label{fig:partition}
\end{figure*}

\textit{On the other hand, human-provided annotations are not always better than MLLM-generated ones.} For certain categories, models trained on MLLM-annotated data surprisingly outperform those trained on human-annotated data. Figure \ref{fig:top_10_ap} illustrates the top-10 categories on O365, ranked by the performance difference between models trained on MLLM-annotated data and those trained on human-annotated data. We observe that for many categories, models trained on MLLM-annotated data significantly outperform those trained on human-annotated data. 
This is primarily because the annotation task in O365 is substantially more challenging than in COCO 2014, as it includes many hard-to-recognize categories, leading to much more manual annotation error compared to COCO 2014. These results suggest that using MLLMs to replace human annotators not only reduces the cost of manual annotation significantly, but also offers certain advantages in annotation quality compared to human annotations. 
Unlike human annotators, whose performance may suffer from inattention or fatigue, MLLMs provide consistent annotations free from such human-induced variability. This further highlights the potential of MLLMs as a scalable and reliable alternative to human annotators.

\begin{figure*}[!t]
    \centering
    \subfigure{
    \centering
    \includegraphics[width=0.6\linewidth]{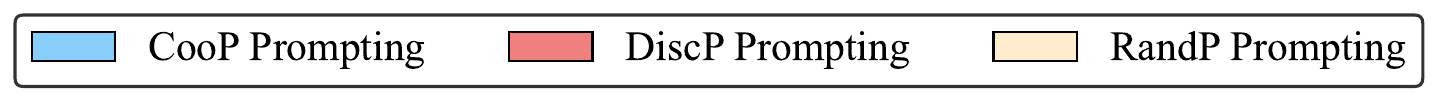}
    }
    \vspace{-0.2cm}

    \subfigure{
        \includegraphics[width=0.3\textwidth]{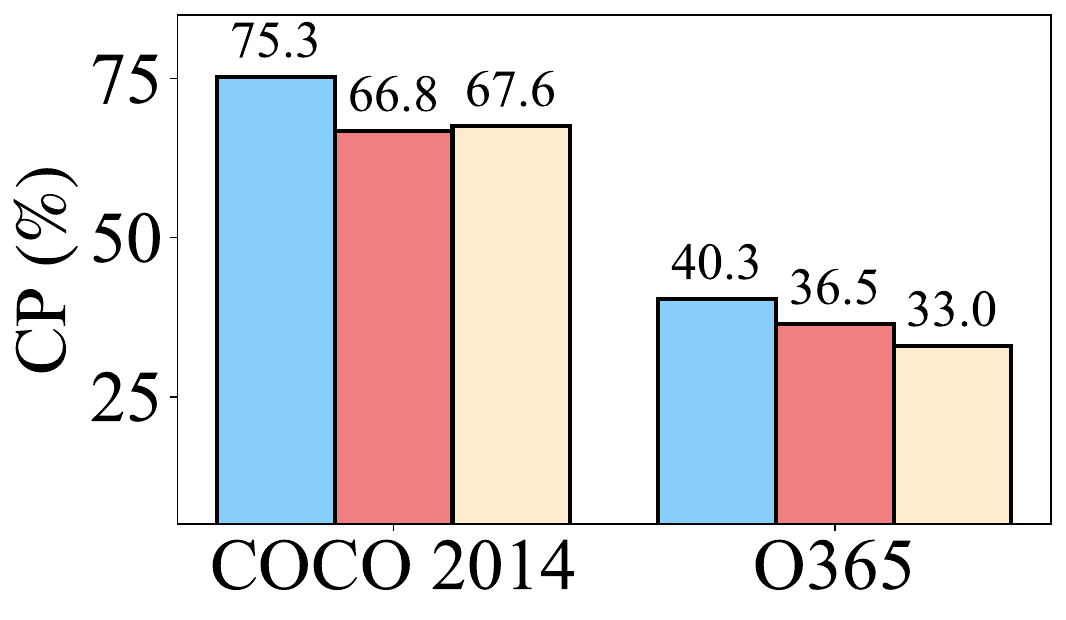}
    }
    \subfigure{
        \includegraphics[width=0.3\textwidth]{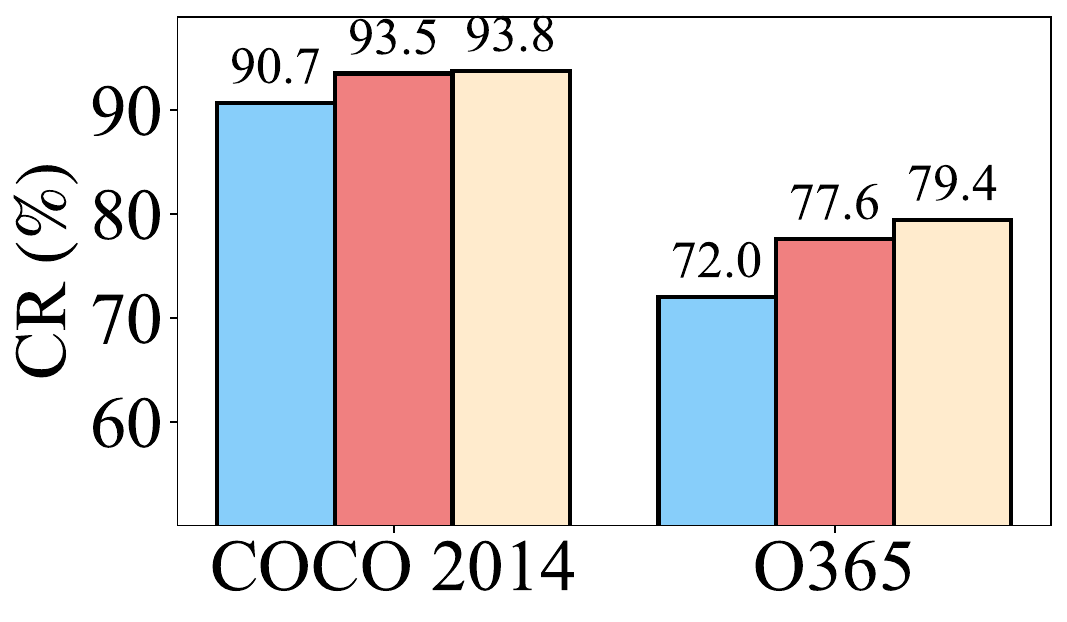}
    }
    \subfigure{
        \includegraphics[width=0.3\textwidth]{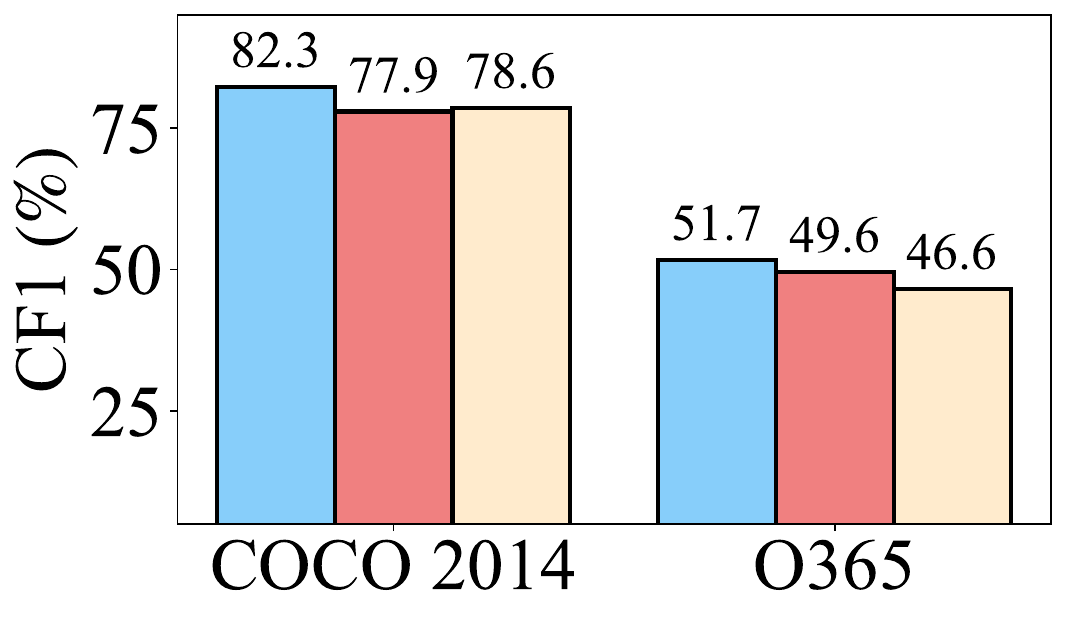}
    }
    \caption{Comparison of three prompting strategies on COCO and Objects365.}
    \label{fig:partition}
\end{figure*}

\section{The TagLLM Framework}

\subsection{The Two-Stage Pipeline}
In this section, we introduce our TagLLM framework, which is designed to generate human-comparable annotations. The framework follows a two-stage design. In the first stage, we use MOP to generate a candidate label set. In the second stage, we use BP to effectively refine the candidate labels. This framework leverages two complementary prompting methods. By combining their respective strengths, it provides a practical and well-motivated annotation pipeline. We highlight the advantages of this framework from two perspectives.
% \begin{itemize}

\textbf{Efficiency}. MOP is used to efficiently generate a compact candidate label set from the original label vocabulary. This substantially reduces the number of binary queries required in the verification stage, thereby improving the overall annotation efficiency. To show how MOP improves annotation efficiency, Figure \ref{fig:mop} illustrates the number of candidate labels produced by MOP and the number of positive labels retained, together with the original number of true labels. After applying MOP, the number of candidate labels drops dramatically, by nearly 20× on both datasets. 
This reduction directly translates into far fewer binary verification queries in the BP stage, thereby substantially lowering the overall annotation cost. 

\textbf{Effectiveness}. The two annotation methods exhibit difference annotation preference.  The combination helps identify more false positive labels, leading to higher annotation quality. To verify that the two-stage pipeline improves annotation quality, Figure \ref{fig:coco_f1} and \ref{fig:o365_f1} present per-class F1 scatter plots comparing the two-stage method with BP. Each point corresponds to a class, and its size is proportional to the number of positive labels for that class. The diagonal line indicates equal performance between the two methods. Points above the diagonal mean the two-stage method outperforms BP, while points below indicate the opposite. The two-stage method outperforms BP for nearly all classes, indicating that the proposed two-stage pipeline can effectively improve annotation quality.
% \end{itemize}

The two-stage framework is only the overall pipeline, and many components within it can be further refined. Below, we present stage-specific improvements for the two stages, respectively.

% As shown in Figure \ref{fig:framework}, the framework mainly consisting of three components: (i) Prompts to candidates method , which uses the group-wise prompting and ensemble techniques to generate candidate label sets; (ii) LVLM-driven disambiguation, which develops the concept-aligned disambiguation strategy to identify true positives from the candidate sets; (iii) Human-assisted calibration, which is an optional choice to refine the candidate label set based on human feedback. 

\subsection{Stage 1: Candidates Generation via Divide-and-Conquer Prompting}
\label{sec:p2c}

% As discussed in Section \ref{sec:prompt}, \textit{binary} prompting typically yields high precision and recall. However, they require $q$ inferences per image, which is substantially more than single inference needed by \textit{multi-option} prompting. By rough estimation, \textit{binary} prompting takes at least ten times longer than \textit{multi-option} prompting. For example, on COCO 2014 with Qwen2.5-VL 7B, \textit{binary} prompting took over 153 hours, whereas \textit{multi-option} prompting required only 10 hours. This resource gap increases with the number of categories and often makes \textit{binary} prompting impractical when the label space comprises hundreds or thousands of categories. This observation motivates a two-stage strategy: first apply \textit{multi-option} prompts to obtain a candidate set, and then apply \textit{binary} prompts to disambiguate and prune false positives. We refer to the first step as \textsc{P2C} (Prompts to Candidates). 
% The approach leverages a division of labor between coverage and verification. 
% \textit{Multi-option} prompting generates a broad candidate set at low cost, thereby reducing the search space. \textit{Binary} prompting then confirms category presence with high precision, pruning spurious labels. The combination attains high recall and high precision with improved cost-effectiveness.

% \textbf{Divide-and-Conquer Prompting}\quad 

In the first stage, the task is to use MOP to generate a candidate label set. When the number of classes is large, MOP faces an additional challenge: the prompt containing all candidate classes becomes long, which increases the risk of hallucinations and omissions. To address this, we adopt a Divide-and-Conquer Prompting (DCP) strategy: partition the label space into multiple groups, perform inference for each group separately, and then merge the group-wise annotations. We consider three partitioning strategies: \textit{Co-occurrence} Partition (CooP), which groups classes that frequently appear together, \textit{Disco-occurrence} Partition (DiscP), which groups categories that rarely appear together, and \textit{Random} Partition (RandP), which groups classes randomly. Given that labels are unknown, we cannot estimate the label co-occurrence probabilities. Instead, We partition the labels into co-occurring and disco-occurring groups by querying ChatGPT. The prompts are provided in Appendix \ref{app:prompt}. Figure \ref{fig:partition} presents annotation performance of three prompting strategies using Qwen3-VL on COCO 2014 and O365. Surprisingly, CooP significantly outperforms DiscP and RandP in precision, at the cost of only a slight decrease in recall. As a result, CooP also achieves substantially higher F1 scores than the other two strategies. One possible explanation is that CooP places classes that are likely to co-occur into the same group. The within-group competition encourages only high-confidence labels, which are more competitive, to be selected, thereby improving annotation precision. We adopt CooP prompting in the first stage to generate the candidate label set.

\begin{table*}[t]
  \centering
  \small
  \caption{The results of annotation quality (OP, OR, OF1, CP, CR, CF1) and model performance (mAP) on COCO 2014.}
    \begin{tabular}{l|c|ccc|ccc|c}
    \toprule
             & \multicolumn{1}{l|}{GPU Time} & OP    & OR    & OF1   & CP    & CR    & CF1   & mAP \\
    \midrule
    NXTP     &        4    & 62.78  & 55.94  & 59.16  & 57.89  & 50.43  & 53.91  & 57.38  \\
    % \midrule
    CLIP     &      $<1$      & 59.30  & 59.20  & 59.25  & 65.00  & 61.65  & 63.28  & 66.85  \\
    % \midrule
    TagCLIP  &        4    & 69.06  & 68.74  & 68.90  & 67.98  & 65.29  & 66.61  & 73.61  \\
    % \midrule
    CaSED    &          3    & 86.22  & 23.85  & 37.36  & 83.30  & 28.70  & 42.69  & 54.42  \\
    % \midrule
    % RAM   &       &       & 89.24  & 55.27  & 68.26  & 89.92  & 62.14  & 73.49  & 79.48  \\
    % \midrule
    RAM++     &          $<1$    & 89.23  & 55.08  & 68.12  & 89.53  & 61.66  & 73.03  & 79.05  \\
    \midrule
             & \multicolumn{8}{c}{Qwen3-VL-8B-Instruct} \\
    \midrule
    MOP      &         12    & 65.63  & 94.75  & 77.54  & 67.59  & 93.81  & 78.57  & 78.80  \\
    % \midrule
    BP    &        166    & 75.77  & 91.68  & 82.97  & 77.11  & 90.28  & 83.18  & 80.76  \\
    \midrule
    TagLLM &        16   & 83.91  & 88.76  & 86.27  & 85.25  & 87.34  & 86.28  & 82.66  \\
    \midrule
          & \multicolumn{8}{c}{Qwen3-VL-30B-A3B-Instruct} \\
    \midrule
    MOP   &         40    & 71.16 & 94.27 & 81.1  & 71.47 & 93.96 & 81.18 & 79.70   \\
    BP    &         600    & 77.13 & 90.75 & 83.39 & 79.86 & 90.37 & 84.79 & 80.52  \\
    \midrule
    TagLLM &       66    & 87.84 & 88.01 & 87.92 & 87.15 & 87.71 & 87.43 & 82.76  \\
    \midrule
    Human Annotator &   --    &   --    &   --    &   --    &   --    &   --    &   --    & 83.26  \\
    \bottomrule
    \end{tabular}%
  \label{tab:coco}%
\end{table*}%

\subsection{Stage 2: Label Refinement via Concept-Aligned Disambiguation}

\label{sec:cad}

% Candidate label disambiguation can be approached from three main directions: (i) LVLM-driven methods, which refine candidate labels through prompt design or LVLMs fine-tuning; (ii) Model-based methods, which train models on candidate labels and rely on their generalization capabilities to identify false positive labels; and (iii) human-assisted methods, which incorporate human feedback to refine candidate labels. Most existing PML studies focus on model-based methods \citep{xie2018partial,zhang2020partial,yang2024noisy}. Although these methods benefit from utilizing the useful information hidden in the candidate sets, they are also prone to overfitting to false positives, which can degrade disambiguation performance. In the proposed framework, we focus on LVLM-driven and human-assisted disambiguation.

In the second stage, BP is used to perform binary verification for each candidate label, thereby refining the candidate set. 
False positive labels in the candidate set are typically attributed to object hallucination \citep{li2023evaluating,leng2024mitigating} (or category hallucination) in MLLMs, where the model incorrectly predicts the presence of objects that are not actually present in the image. Our analysis reveals that object hallucination is caused not only by the inherent recognition limitations of MLLMs, but also by several external factors. An important factor is the misalignment between category names and intended semantic concepts, especially for those uncommon or ambiguous classes. We summarize three types of category-concept misalignments as follows:
\begin{itemize}
    \item \textbf{Sense ambiguity} refers to cases where a category name is semantically unclear and may carry multiple interpretations. 
    % potentially referring to distinct types of objects. 
    % For example, the category name \textit{orange}, which typically refers to the fruit but can also denote the color. 
    %and \textit{apple}, which usually refers to the fruit, but may also be interpreted as the technology brand. 
    % These semantic ambiguities frequently cause the model to assign a large number of false positives, ultimately reducing the overall precision.
    \item \textbf{Hypernym overreach} refers to cases where a category name covers an excessively wide range of meanings.
    % often extending beyond what the user originally intended to refer to. 
    % For example, in the COCO 2014 and O365 datasets, the category name \textit{tie} typically refers to a \textit{necktie}. However, the word \textit{tie} can also broadly denote any knot-like object, leading to unintended over-generalization in LVLM annotation.
    \item \textbf{Misnomer/underspecification} refers to the use of category names that fail to precisely capture the intended semantic meaning. 
    % making them unable to properly represent what the user originally intended to annotate. 
    % For example, in O365, the category names \textit{mask} and \textit{marker} are used to denote \textit{face mask} and \textit{marker pen}, respectively. 
\end{itemize}
More detailed definitions and examples are provided in Appendix~\ref{app:misalign}.

These three types of category-concept misalignment often cause MLLMs to generate excessive false positives, thereby reducing annotation precision and adversely affecting the performance of models trained on MLLM-annotated data. To mitigate this issue, we propose a Concept-Aligned Disambiguation (CAD) strategy, which aims to leverage MLLMs to refine the candidate label set by designing prompts that are unambiguous and semantically clear. We summarize the three steps as follows \footnote{Detailed information can be found in Appendix~\ref{app:cad}}:
\begin{itemize}
    \item For each category $C_k$, we ask ChatGPT-4o to identify the super-category for each category and incorporate the following sentence into the annotation prompt: \prompt{$C_k$ is a type of $M_k$.}.

    \item For each category $C_k$, we ask ChatGPT-4o to return up to five categories from the entire category set whose visual appearance is most similar to $C_k$. We define the resulting set of similar categories as $\{S_j\}_{j=1}^{o_k}$, where $o_k$ denotes the number of similar categories. We then append the following sentence to the annotation prompt: \prompt{$C_k$ does not refer to $S_1$, $S_2$, \ldots, or $S_{o_k}$.}.

    \item For each category $C_k$, we prompt ChatGPT-4o to return a set of descriptions $D_k$ that can help eliminate its ambiguity based on both semantic relevance and common usage frequency. Then, we substitute the original category name $C_k$ in the annotation prompt with $D_k$, yielding the refined prompt: \prompt{Carefully examine the image and decide if it contains a $D_k$. Answer with only yes or no.}.
\end{itemize}
% For each of the steps described above, once the final annotation prompt is generated, we manually review it to ensure its correctness, and make necessary adjustments if any part is found to be inappropriate or unclear.

% \subsection{Human-Assisted Calibration}

% An alternative approach involves leveraging human feedback to refine the candidate label set. Unlike traditional manual annotation, which requires annotating the entire label set, our proposed Human-Assisted Calibration (HAC) only requires humans to verify a candidate label set that is orders of magnitude smaller. Specifically, using the P2C method proposed in Section \ref{sec:p2c}, the average number of candidate labels that need to be checked per image is 5.66 for COCO 2014 and 18.16 for O365. Furthermore, if the candidate sets are further refined by the CAD method introduced in Section \ref{sec:cad}, the number of labels that require human verification is reduced to 2.96 for COCO 2014 and 6.86 for O365. In Section \ref{sec:experiments}, we will present experimental results to demonstrate both model performance and the scale of cost reduction in human annotation.

\begin{table*}[t]
  \centering
  \small
  \caption{The results of annotation quality (OP, OR, OF1, CP, CR, CF1) and model performance (mAP) on O365.}
    \begin{tabular}{l|c|ccc|ccc|c}
    \toprule
          & GPU Time & OP    & OR    & OF1   & CP    & CR    & CF1   & mAP \\
    \midrule
    NXTP  &          4    & 34.71  & 25.11  & 29.14  & 41.72  & 18.52  & 25.65  & 23.81  \\
    % \midrule
    CLIP  &            $<1$    & 39.60  & 40.87  & 40.22  & 31.90  & 29.94  & 30.89  & 27.75  \\
    % \midrule
    TagCLIP &         9    & 51.71  & 50.53  & 51.11  & 38.01  & 31.99  & 34.74  & 33.34  \\
    % \midrule
    CaSED &            3    & 63.84  & 6.41  & 11.65  & 48.02  & 11.12  & 18.05  & 22.08  \\
    % \midrule
    % RAM   &       &       & 66.15  & 23.70  & 34.90  & 55.95  & 31.68  & 40.45  & 40.09  \\
    % \midrule
    RAM++  &            $<1$    & 66.11  & 23.54  & 34.72  & 54.71  & 31.25  & 39.78  & 39.43  \\
    \midrule
          & \multicolumn{8}{c}{Qwen3-VL-8B-Instruct} \\
    \midrule
    MOP   &            30    & 39.50  & 85.38  & 54.01  & 32.98  & 79.42  & 46.61  & 43.25  \\
    % \midrule
    BP    &            650    & 45.57  & 79.63  & 57.97  & 37.02  & 75.67  & 49.71  & 43.45  \\
    \midrule
    TagLLM &           66    & \multicolumn{1}{c}{63.38 } & \multicolumn{1}{c}{70.83 } & \multicolumn{1}{c}{66.90 } & 53.48  & 65.57  & 58.91  & 46.50  \\
    \midrule
          & \multicolumn{8}{c}{Qwen3-VL-30B-A3B-Instruct} \\
    \midrule
    MOP   &           92    & 39.98 & 84.28 & 54.23 & 32.04 & 82.64 & 46.18 & 43.50  \\
    BP    &           2780   & 53.64  & 73.53  & 62.03  & 42.15  & 73.90  & 53.68  & 44.88  \\
    \midrule
    TagLLM &          242   & 67.59 & 67.91 & 67.75 & 54.99 & 67.07 & 60.43 & 47.13  \\
    \midrule
    Human Annotator  &   --    &  --     &   --    &    --   &   --    &   --    &   --    & 48.58  \\
    \bottomrule
    \end{tabular}%
  \label{tab:o365}%
\end{table*}%

\begin{figure*}[t]
    \centering
    \subfigure{
        \includegraphics[width=0.6\textwidth]{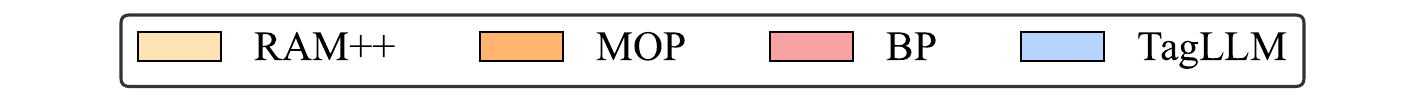}
    }\vspace{-0.4cm}
    \subfigure{
        \includegraphics[width=0.95\textwidth]{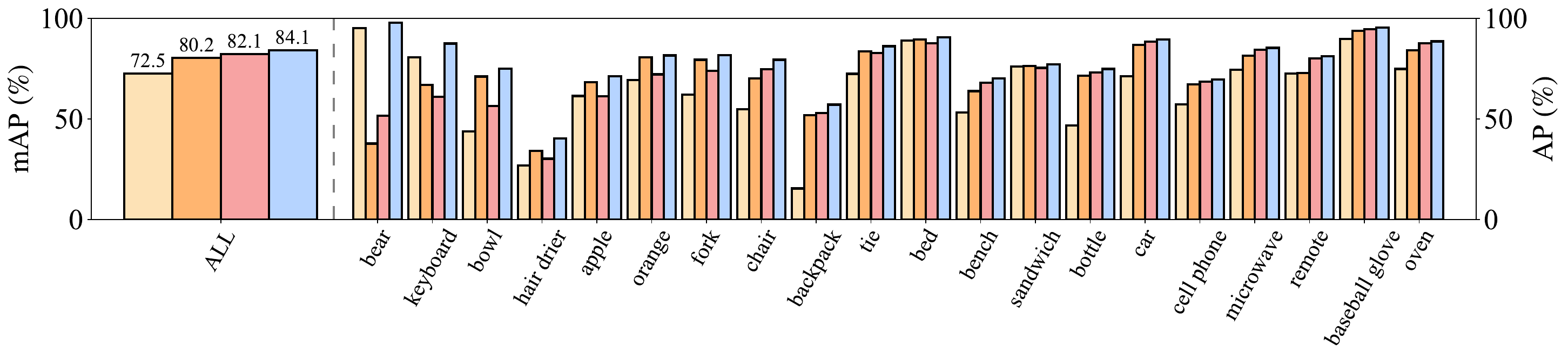}
    }
    \caption{The performance on COCO 2014 validation set of ResNet-101 trained on annotations generated by difference methods on COCO 2017 unlabeled set.}
    \label{fig:coco2017}
    \vspace{-1em}
\end{figure*}

\section{Experiments}
\label{sec:experiments}

In this section, we first present the main experimental results, followed by further analyses. Due to space limit, detailed experimental settings are provided in the Appendix \ref{sec:app_setting}.

\subsection{Experimental Settings}

\paragraph{Datasets} To evaluate the performance of the proposed method, we perform experiments on three benchmark datasets, including MS-COCO 2014 (COCO 2014), MS-COCO 2017 (COCO 2017), and Objects365 (O365). We perform several pre-processing steps on all datasets; detailed dataset descriptions and pre-processing protocols are provided in Appendix \ref{sec:app_setting}. For COCO 2017 in particular, we use the unlabeled split as a practical application scenario for our method, which contains approximately 123 $k$ images without manual annotations. Since ground-truth labels are unavailable, we cannot directly score annotation quality. Instead, we train models on the annotations produced by each method and assess annotation quality via their performance on the COCO 2014 validation set.

\paragraph{Evaluation Metrics} We adopt two types of evaluation metrics to assess annotation quality and model performance separately. For annotation quality, we report Overall Precision (OP), Overall Recall (OR), Overall F1-score (OF1), per-Class Precision (CP), per-Class Recall (CR), and per-Class F1-score (CF1). To evaluate model performance, we use mean Average Precision (mAP). Detailed definitions of these metrics can be found in Appendix \ref{sec:app_setting}. We also report the GPU time required for annotation to assess the efficiency of each method.

\paragraph{Comparing Methods} To validate the effectiveness of TagLLM, besides the baselines BP and MOP, we compare it against multiple state-of-the-art image tagging methods: CLIP \citep{radford2021learning}, TagCLIP \citep{lin2024tagclip}, RAM++ \citep{huang2023open}, CaSED \citep{conti2023vocabularyfree}, and NXTP \citep{yue2024object} (see Appendix \ref{sec:app_setting} for detailed information). We also compare against human annotators (see Appendix \ref{sec:app_setting} for detailed information).

\paragraph{Implementation} For downstream model training, we use standard ResNet101 \citep{he2016deep} with a learning rate of $0.0001$ and a batch size of 128. We use the AdamW \citep{LoshchilovH19} optimizer and the OneCycle Scheduler. Furthermore, we perform exponential moving average (EMA) \citep{tarvainen2017mean} for the model parameters with a decay of 0.9997. We use the ASL loss \citep{ridnik2021asymmetric} as the base loss function.

\begin{table*}[t]
  \centering
  \caption{Ablation Studies on COCO 2014 and O365.}
    \begin{tabular}{ccccc|ccc|ccc}
    \toprule
    \multicolumn{5}{c|}{Method}          & \multicolumn{3}{c|}{COCO 2014} & \multicolumn{3}{c}{O365} \\
    \midrule
    MOP & BP & Two-Stage & DCP & CAD & OF1   & CF1   & mAP   & OF1   & CF1   & mAP \\
    \midrule
    \checkmark     &       &       &       &       & 77.54  & 78.57  & 78.80  & 54.01  & 46.61  & 43.25  \\
    \midrule
          & \checkmark     &       &       &       & 82.97  & 83.18  & 80.76  & 57.97  & 49.71  & 43.45  \\
    \midrule
    \checkmark     & \checkmark     & \checkmark     &       &       & 84.82  & 84.91  & 81.58  & 63.05 & 54.86 & 45.22  \\
    \midrule
    \checkmark     & \checkmark     & \checkmark    & \checkmark     &       & 85.56  & 85.68  & 81.71  & 65.29 & 57.57 & 46.19  \\
    \midrule
    \checkmark     & \checkmark     & \checkmark     & \checkmark    & \checkmark     & 87.92  & 87.43  & 82.76  & 66.90  & 58.91  & 46.50  \\
    \bottomrule
    \end{tabular}%
  \label{tab:ablation}%
\end{table*}%

\begin{figure*}[!tb]
    \centering
    \includegraphics[width=0.9\textwidth]{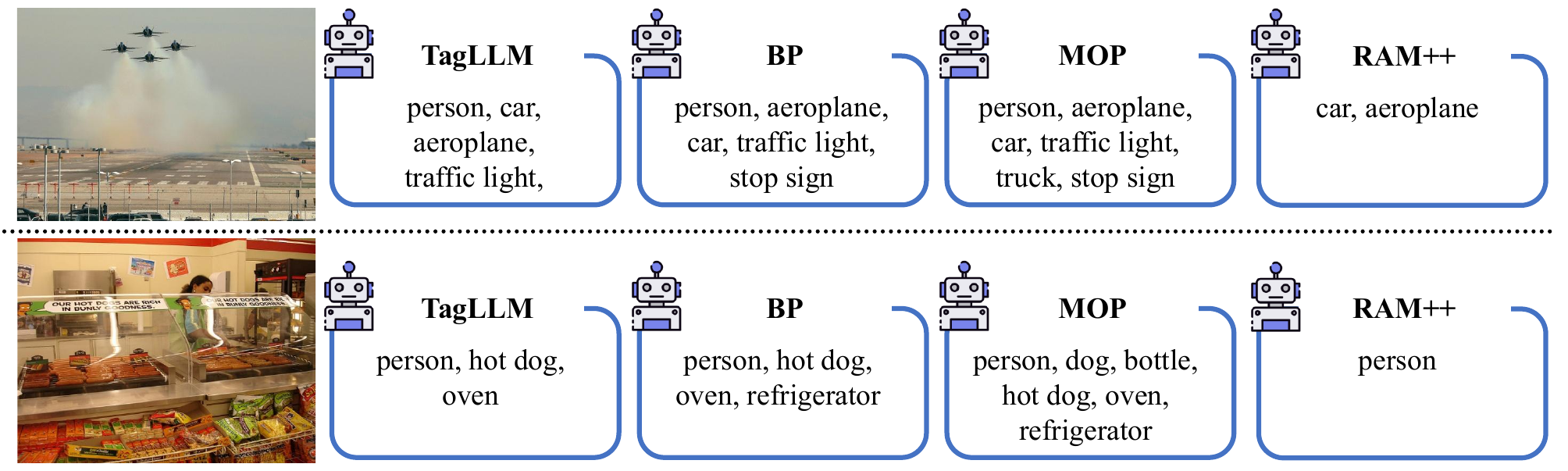}
    \caption{Visualization of annotations generated by different methods on COCO 2017.}
    \label{fig:case_study}
    \vspace{-1em}
\end{figure*}

\subsection{Comparison Results}

Table \ref{tab:coco} and Table \ref{tab:o365} report comparisons of TagLLM with the state-of-the-art methods and human annotations in terms of both annotation quality and downstream model performance on COCO 2014 and O365. We can see that our proposed TagLLM consistently outperforms the other methods in terms of both annotation quality and model performance. The resulting performance is only about 0.5\% and 1.4\% lower in mAP compared to using human annotations on COCO 2014 and O365, respectively. Compared with the optimal baseline, TagLLM reduces the downstream performance gap between model-generated and human annotations by 80\% on COCO 2014 (from 2.5\% to 0.5\%) and by 60\% on O365 (from 3.7\% to 1.4\%).
% Incorporating human feedback via HAC further improves annotation quality and model performance. 
% Compared to manual annotation, using the proposed methods, TagLLM and P2C, significantly reduces annotation costs. On COCO 2014, the number of labels requiring annotation drops to 5.66 and 2.96, reducing the cost by approximately 14× and 27×. On O365, the numbers are 18.16 and 6.86, corresponding to 14× and 36× reductions. 
We also report the GPU time required for annotation with each method. Although the discriminative methods require the least time, they obtain unfavorable annotation performance. The proposed TagLLM is markedly more efficient than the other methods. 
% From the tables, we can see: i) Our proposed \textsc{LVLMAnt} consistently outperforms the other methods in terms of both annotation quality and model performance. ii) Our method achieves performance comparable to that of human annotators on both datasets. When training models using the labels generated by our method, the resulting performance is only about 0.5\% and 1.6\% lower in mAP compared to using human annotations on COCO 2014 and O365, respectively. iii) Incorporating human feedback via HAC further improves annotation quality and model performance. Compared to manual annotation, using LVLMs (\textsc{LVLMAnt} or P2C) significantly reduces annotation costs. On COCO 2014, the number of labels requiring annotation drops to 5.66 and 2.96, reducing the cost by approximately 14× and 27×. On O365, the numbers are 18.16 and 6.86, corresponding to 14× and 36× reductions. We also report the GPU time required for annotation with each method. Although the discriminative methods (RAM and RAM++) require the least time, they obtain unfavorable annotation performance. The proposed \textsc{LVLMAnt} is markedly more efficient than the baseline prompting methods. 

% Table generated by Excel2LaTeX from sheet 'Sheet1'
% Table generated by Excel2LaTeX from sheet 'Sheet1'

To assess the practical effectiveness of TagLLM, we conduct experiments on the COCO 2017 unlabeled image set and report performance on the COCO 2014 validation set. Figure \ref{fig:coco2017} shows the overall mAP and the per class AP for all methods. Due to space limit, we display only the 20 categories where TagLLM achieves the largest gains over the baselines, in order to analyze the sources of its advantages. TagLLM achieves the best performance and significantly outperforms the comparing methods. TagLLM substantially improves disambiguation for ambiguous categories, such as \textit{orange}, \textit{tie}, and \textit{apple}, yielding high-quality annotations.  

% These experimental results convincingly demonstrate that our proposed annotation framework can serve as an effective and efficient substitute for human annotators in neural image tagging.

\subsection{Further Studies}

To understand why the proposed method achieves high-quality annotations, we conduct ablation studies on each component of the pipeline. Table \ref{tab:ablation} presents the results of these experiments on the COCO2014 and O365 datasets. As shown in the table, each component contributes to the final performance. The two-stage pipeline provides a generally applicable gain, while DCP and CAD deliver more pronounced improvements on specific datasets. 

% Furthermore, Table 4 presents the reduction in annotation cost (in terms of the number of labels needed to be annotated) when using candidate sets generated by the proposed method, along with the resulting model performance.

% Table generated by Excel2LaTeX from sheet 'Sheet1'

% \begin{table*}[!t]
%     \centering
%     \small
%     \caption{Abalation Studies on O365. S.-C. denotes super-category.}
%     \label{tb:ablation}
%     \begin{tabular}{cccc|ccc|ccc|c}
%         \toprule
%         P2C         & S.-C.      & Similarity & Description  & OP       & OR      & OF1   & CP    & CR    & CF1   & mAP \\
%         \midrule
%         \checkmark  &            &            &              & 28.27 	&  84.87  &	42.41 & 18.54 &	87.06 &	27.66 & 39.63   \\
%         \checkmark  & \checkmark &            &              & 56.22 	&  72.12  & 63.19 & 42.24 &	71.68 &	53.15 & 46.08    \\
%         \checkmark  & \checkmark & \checkmark &              & 56.72 	&  70.82  &	62.99 & 43.15 &	71.12 &	53.71 & 46.29   \\
%         \checkmark  & \checkmark & \checkmark & \checkmark   & 60.43 	&  68.52  & 64.22 & 46.99 &	67.79 &	55.50 & 46.99   \\
%         \bottomrule
%     \end{tabular}
% \end{table*}

% \subsection{Case Studies}

Finally, we present qualitative annotations on COCO 2017 produced by different methods. From Figure \ref{fig:case_study}, RAM++ yields precise annotations but often misses many labels. Compared with the baselines, TagLLM achieves higher precision and broader coverage. These gains arise from the proposed two-stage pipeline, whose structured annotation paradigm substantially improves annotation quality.

\section{Related Works}

Recent studies have explored leveraging LLMs for data annotation.
% , either as standalone annotators or as assistants to human annotators. 
\citet{wang-etal-2021-want-reduce} made the first attempt to leverage GPT-3 as a low-cost data annotator for natural language processing (NLP) tasks. \citep{ding-etal-2023-gpt} evaluated GPT-3 for annotating and augmenting data in classification and token-level tasks. Several studies have found that, on certain tasks, LLMs can surpass crowdsourced annotators in annotation quality \citep{pnas,he-etal-2024-annollm}. \citet{wang2024} developed an annotation framework to leverage the strengths of both LLMs and humans to ensure the accuracy and reliability of annotations. \citet{kim-etal-2024-meganno} developed MEGAnno+ system to facilitate human-LLM collaboration through efficient LLM annotation and selective human verification. Despite the great successes that these methods achieved, they mainly focused on NLP tasks and cannot be directly applied to image tagging tasks. Recognize Anything Model (RAM) \citep{zhang2024recognize} and its improved version RAM++ \citep{huang2023open} were proposed for image tagging tasks. However, their annotation quality still lags behind that of human annotators and therefore cannot replace human annotation. Some studies reduce annotation cost by training models on a small number of labeled images \citep{xie2023classdistributionaware,xiao2024dual,kou2025rankmatch}.

\section{Conclusion}

% This paper studies the problem of image tagging by leveraging large vision-language models (LVLMs) to reduce reliance on manual annotations. 
This paper proposes an automated image tagging framework, TagLLM, which integrates the group-wise prompting method to generate diverse candidate labels and employs the concept alignment mechanism to resolve semantic ambiguities through iterative refinement. The group-wise prompting approach ensures broad coverage of potential labels, whereas the alignment strategy dynamically calibrates category semantics using advanced language models to mitigate label-concept mismatches. Extensive experiments on benchmark datasets demonstrate the effectiveness of TagLLM in achieving high-quality annotations comparable to human efforts, with significant reductions in annotation costs. A limitation of this work is that we only evaluate TagLLM on natural image datasets. In future work, we plan to extend our evaluation to domain-specific datasets, such as those in the fine-grained image tagging domain.

% Acknowledgements should only appear in the accepted version.
% \section*{Acknowledgements}

% \textbf{Do not} include acknowledgements in the initial version of the paper
% submitted for blind review.

% If a paper is accepted, the final camera-ready version can (and usually should)
% include acknowledgements.  Such acknowledgements should be placed at the end of
% the section, in an unnumbered section that does not count towards the paper
% page limit. Typically, this will include thanks to reviewers who gave useful
% comments, to colleagues who contributed to the ideas, and to funding agencies
% and corporate sponsors that provided financial support.

\section*{Impact Statement}

This paper presents work whose goal is to advance the MLLM annotation for image tagging. There are many potential societal consequences of our work, none
which we feel must be specifically highlighted here.

% In the unusual situation where you want a paper to appear in the
% references without citing it in the main text, use \nocite

\nocite{langley00}

\bibliography{example_paper}
\bibliographystyle{icml2026}

%%%%%%%%%%%%%%%%%%%%%%%%%%%%%%%%%%%%%%%%%%%%%%%%%%%%%%%%%%%%%%%%%%%%%%%%%%%%%%%
%%%%%%%%%%%%%%%%%%%%%%%%%%%%%%%%%%%%%%%%%%%%%%%%%%%%%%%%%%%%%%%%%%%%%%%%%%%%%%%
% APPENDIX
%%%%%%%%%%%%%%%%%%%%%%%%%%%%%%%%%%%%%%%%%%%%%%%%%%%%%%%%%%%%%%%%%%%%%%%%%%%%%%%
%%%%%%%%%%%%%%%%%%%%%%%%%%%%%%%%%%%%%%%%%%%%%%%%%%%%%%%%%%%%%%%%%%%%%%%%%%%%%%%
\newpage
\appendix
\onecolumn
\section{Prompt Design}
\label{app:prompt}

Table \ref{tab:prompt} shows the prompt templates for binary prompting and multi-option prompting. Table \ref{tab:cp_dp} shows the the prompt templates for CooP and DiscP.

\begin{table}[htbp]
  \centering
  \caption{Prompt templates for binary prompting and multi-option prompting}
    \begin{tabular}{c|p{18em}|p{18em}}
    \toprule
              & \multicolumn{1}{c|}{Simple} & \multicolumn{1}{c}{Carefully-Designed} \\
    \midrule
    \multicolumn{1}{l|}{Binary Prompt} & Is there a \{object\} in this image?. Answer with only yes or no. & Carefully examine the image and decide if it contains a \{object\}. Answer with only yes or no. \\
    \midrule
    Multi-Option Prompt & What objects are in this image? Candidates: \{cls\_list\}. From this list, output only the names of the objects that are present, separated by commas. Do not include any object that is not in the candidate list. If none of the candidate objects are present, output exactly NO. & Carefully examine the image and decide which of the following candidate objects are present in the image. Candidates: \{cls\_list\}. From this list, output only the names of the objects that are present, separated by commas. Do not include any object that is not in the candidate list. If none of the candidate objects are present, output exactly NO. \\
    \bottomrule
    \end{tabular}%
  \label{tab:prompt}%
\end{table}%

\begin{table}[htbp]
  \centering
  \caption{Prompt templates for CooP and DiscP.}
    \begin{tabular}{p{20em}|p{20em}}
    \toprule
    \multicolumn{1}{c|}{The Pompt for CooP} & \multicolumn{1}{c}{The Pompt for DiscP} \\
    \midrule
    Based on the co-occurrence relationships among the categories, please divide these n categories into m groups, ensuring that the categories within each group frequently co-occur. & Based on the co-occurrence relationships among the categories, please divide these n categories into m groups, ensuring that the categories within each group do not co-occur. \\
    \bottomrule
    \end{tabular}%
  \label{tab:cp_dp}%
\end{table}%

\section{Details of Category-Concept Misalignment}
\label{app:misalign}

Table \ref{tab:misalign} demonstrates the definitions and examples of three types of category-concept misalignment.

\section{Details of Concept-Aligned Disambiguation}
\label{app:cad}
We provide detailed descriptions of the three steps in CAD:
\begin{itemize}
    \item For each category $C_k$, we ask ChatGPT-4o to identify the super-category for each category by providing the following prompt: \prompt{Which super-category does $C_k$ belong to? For example, an apple is a type of fruit, and a car is a type of vehicle.} After obtaining the super-category for each category $C_k$, denoted as $M_k$, we incorporate the following sentence into the annotation prompt: \prompt{$C_k$ is a type of $M_k$.}.

    \item For each category $C_k$, we ask ChatGPT-4o to return up to five categories from the entire category set whose visual appearance is most similar to $C_k$. We define the resulting set of similar categories as $\{S_j\}_{j=1}^{o_k}$, where $o_k$ denotes the number of similar categories. We then append the following sentence to the annotation prompt: \prompt{$C_k$ does not refer to $S_1$, $S_2$, \ldots, or $S_{o_k}$.}.

    \item For each category $C_k$, we prompt ChatGPT-4o with the question: \prompt{Does a category name $C_k$ have multiple meanings? If so, please provide several concise phrases that can help eliminate its ambiguity.}. ChatGPT-4o will return a set of phrases based on both semantic relevance and common usage frequency. In real-world scenarios, users can flexibly choose suitable phrases to compose the description $D_k$ according to their preferences or requirements. For consistency in our experiments, we constructed $D_k$ using the top three phrases ranked by ChatGPT-4o. Then, we substitute the original category name $C_k$ in the annotation prompt with $D_k$, yielding the refined prompt: \prompt{Carefully examine the image and decide if it contains a $D_k$. Answer with only yes or no.}.
\end{itemize}

\begin{table}[htbp]
  \centering
  \caption{Definitions and examples of three types of category-concept misalignment.}
    \begin{tabular}{p{6em}|p{18em}|p{18em}}
    \toprule
    & \multicolumn{1}{c|}{Definition} & \multicolumn{1}{c}{Examples} \\
    \midrule
    Sense Ambiguity & A category name is semantically unclear and may carry multiple interpretations, potentially referring to distinct types of objects.  & In COCO 2014, the category name \textit{orange}, which typically refers to the fruit but can also denote the color and \textit{apple}, which usually refers to the fruit, but may also be interpreted as the technology brand.  \\
    \midrule
    Hypernym Overreach & A category name covers an excessively wide range of meanings, often extending beyond what the user originally intended to refer to. & In COCO 2014 and O365, the category name \textit{tie} typically refers to a \textit{necktie}. However, the word \textit{tie} can also broadly denote any knot-like object. \\
    \midrule
    Misnomer/ Underspecification & A category name fails to precisely capture the intended semantic meaning, making them unable to properly represent what the user originally intended to annotate.   & In O365, the category names \textit{mask} and \textit{marker} are used to denote \textit{face mask} and \textit{marker pen}, respectively. \\
    \bottomrule
    \end{tabular}%
  \label{tab:misalign}%
\end{table}%

\begin{table}[t]
\caption{The statics of benchmark datasets used.}
\label{tb:dataset}
\centering
\begin{tabular}{@{}cccccc@{}}
\specialrule{1.5pt}{0pt}{3pt}
\multirow{2}{*}{\textbf{Dataset}} &
\multirow{2}{*}{\textbf{\#   CLS}} &
\multirow{2}{*}{\textbf{\#   Train}} &
\multirow{2}{*}{\textbf{\#   Val}} &
  \multicolumn{2}{c}{\textbf{\#   Labels / Image}} \\ \cmidrule(l){5-6} 
            &     &           &        & \textbf{Min$\sim$Max} & \textbf{Avg} \\ \midrule
COCO-2014            & 80     & 82,081   & 40,137 & 1$\sim$18         & 2.9            \\
Objects365-filtered           & 251    & 104,395  & 78,303 & 1$\sim$30         & 6.3            \\
COCO-2017-unlabeled & 80     & 123,403  & -      & -                 & -              \\ \bottomrule

\end{tabular}
\end{table}

\section{Detailed Experimental Settings}

\label{sec:app_setting}

% COCO 2014 consists of 82,081 training images and 40,137 validation images for 80 classes, with an average of 2.9 labels per image. O365 presents significant annotation challenges, due to its large scale, encompassing 365 categories and more than 1.8 million images. Annotating such a large number of images using large models is highly challenging and time-consuming. To improve efficiency, we randomly sampled a subset of approximately 100,000 images from the full dataset. In addition, we applied a preprocessing step to retain only the labels with more than 100 positive instances. This step is important for accurately assessing the impact of annotation quality. We found that if labels with very few positive samples are kept, their consistently poor performance, regardless of annotation quality, can reduce the apparent influence of annotation quality on overall model performance. In such cases, the presence of extremely rare labels makes it difficult to observe meaningful differences, since the model tends to perform poorly on these labels no matter how well they are annotated. We also use the COCO 2017 unlabeled set as a practical application scenario for our method. It contains approximately 123k images without manual annotations. Given that ground-truth labels are unavailable, we cannot directly score annotation quality. We therefore train models on datasets annotated by each method and assess annotation quality indirectly via their performance on the COCO 2014 validation set. 

\paragraph{Datasets} To evaluate the performance of the proposed method, we perform experiments on three benchmark datasets, including MS-COCO 2014 (COCO 2014), MS-COCO 2017 (COCO 2017), and Objects365 (O365). We perform several pre-processing steps on all datasets. We utilize the original train/validation split from the 2014 version of the COCO \footnote{\url{https://cocodataset.org/##home}} dataset, which contains 82,081 training images and 40,137 validation images for 80 classes, with an average of 2.9 labels per image. O365 \footnote{\url{https://www.objects365.org/overview.html}}  presents significant annotation challenges, due to its large scale, encompassing 365 categories and more than 1.8 million images. Annotating such a large number of images using large models is highly challenging and time-consuming. To improve efficiency, we randomly sampled a subset of approximately 100,000 images from the full dataset. In addition, we applied a preprocessing step to retain only the labels with more than 100 positive instances. This step is important for accurately assessing the impact of annotation quality. We found that if labels with very few positive samples are kept, their consistently poor performance, regardless of annotation quality, can reduce the apparent influence of annotation quality on overall model performance. In such cases, the presence of extremely rare labels makes it difficult to observe meaningful differences, since the model tends to perform poorly on these labels no matter how well they are annotated. This results in 104,395 training images and 78,303 validation images, covering 251 categories, with an average of 6.3 labels per image.
% The COCO 2017 unlabeled dataset contains 123,403 unlabeled images. 
% For COCO 2017 in particular, we use the unlabeled split as a practical application scenario for our method, which contains 123,403 images without manual annotations. Since ground-truth labels are unavailable, we cannot directly score annotation quality. Instead, we train models on the annotations produced by each method and assess annotation quality via their performance on the COCO 2014 validation set. 
The detail characteristics of three benchmark datasets are reported in Table \ref{tb:dataset}. 

\paragraph{Evaluation Metrics} We employs two types of evaluation metrics to assess annotation quality and model performance, respectively. For annotation quality, we use six metrics: Overall Precision (OP), Overall Recall (OR), Overall F1-score (OF1), per-Class Precision (CP), per-Class Recall (CR), and per-Class F1-score (CF1). Specifically, these metrics can be computed as follows:
\begin{equation}\nonumber
\label{eq:cf1_of1}
\text{CF1}=\frac{2\times \text{CP} \times \text{CR}}{\text{CP} + \text{CR}}, \ \ \ \text{OF1}=\frac{2\times \text{OP} \times \text{OR}}{\text{OP} + \text{OR}},
\end{equation}
and
\begin{equation}\nonumber
\begin{aligned}
\label{eq:cp_op_cr_or}
&\text{CP}=\frac{1}{K}\sum_{k}\frac{N_k^{TP}}{N_k^{TP}+N_k^{FP}}, &\text{OP}=\frac{\sum_{k}N_k^{TP}}{\sum_{k}(N_k^{TP}+N_k^{FP})},\\
&\text{CR}=\frac{1}{K}\sum_{k}\frac{N_k^{TP}}{N_k^{TP}+N_k^{FN}}, &\text{OR}=\frac{\sum_{k}N_k^{TP}}{\sum_{k}(N_k^{TP}+N_k^{FN})},
\end{aligned}
\end{equation}
where CP, CR are average per-class precision, recall, and OP, OR are overall precision, recall. According to the confusion matrix, $\{N_k^{TP}, N_k^{FP}, N_k^{TN}, N_k^{FN}\}$ indicate the number of true positive, false positive, true negative, false negative for the $k$-th class. 
% The superiority of our approach is validated by these experimental results.

To evaluate model performance, we use mean Average Precision (mAP) to measure the model trained under various annotation strategies. It is defined as follows:
\begin{equation}\nonumber
    \mathrm{mAP}=\frac{1}{K} \sum_{k=1}^{K} \operatorname{AP}_k ,
    \quad \operatorname{AP}_k = \operatorname{AP}\left(\p_{\cdot k}, \boldsymbol{y}_{\cdot k}\right)=\frac{1}{|\boldsymbol{y}_{\cdot k}|} \sum_{i: y_{i k}=1} \frac{\left|\left\{y_{j k}=1: \p_{j k}>\p_{i k}\right\}\right|}{\operatorname{\textbf{rank}}\left(\p_{\cdot k}\right)_{i}},
\end{equation}
where $\p$ and $\y$ denote the predicted probabilities and ground truth labels, respectively, and mAP is the mean of the average precision (AP) over all classes. The denominator ${\operatorname{\textbf{rank}}\left(\p_{\cdot k}\right)_{i}}=1+|\{j: \p_{jk} > \p_{ik}\}|$ represents the rank of the model output $\p_{ik}$ among all predictions for class $k$, and the numerator ${\left|\left\{y_{j k}=1: \p_{j k}>\p_{i k}\right\}\right|}$ counts the number of true positive predictions that have a higher predicted probability than $\p_{ik}$.

\paragraph{Comparing Methods} To validate the effectiveness of TagLLM, we compare it against the state-of-the-art methods. Below, we provide a detailed introduction to these compared methods.
\begin{itemize}
\item \textbf{CLIP}~\citep{radford2021learning} is a vision-language model that learns a shared embedding space for images and text via contrastive learning on large-scale image–caption pairs, enabling strong zero-shot recognition and retrieval.

\item \textbf{TagCLIP}~\citep{lin2024tagclip} is a training-free local-to-global framework built on a frozen CLIP-ViT that produces high-quality open-vocabulary image labels by combining patch-level scoring, dual-masking attention refinement, and class-wise re-identification.

% \item \textbf{RAM}~\citep{} is a foundational model for image tagging that leverages large-scale automatically parsed tags from image-text pairs. Using a Swin Transformer as the image encoder and a lightweight tag recognition decoder, RAM achieves high-accuracy zero-shot recognition of over 6,400+ common categories. Without relying on manual annotations, the model constructs its training data through semantic text parsing and an automated tag-cleaning engine. It significantly outperforms general-purpose vision-language models like CLIP and BLIP on multiple image tagging benchmarks.

\item \textbf{RAM++}~\citep{huang2023open} is an open-set image tagging model that leverages multi-grained text supervision to recognize both predefined common categories and diverse unseen categories with high accuracy..

\item \textbf{CaSED}~\citep{conti2023vocabularyfree} is a vocabulary-free image tagging model that exploits a pre-trained vision-language model and an external vision-language database to provide labels for images. 

\item \textbf{NXTP}~\citep{yue2024object} reframes image classification as autoregressive next-token prediction with a language decoder, enabling vocabulary-free recognition over a broad textual space.

\end{itemize}

\begin{table}[!t]
\centering
\caption{Category difference comparison: Original dataset category to ($\rightarrow$) RAM category.}
\label{tb:diff}
\begin{tabular}{@{}r c l @{\hspace{2em}} r c l@{}}
\toprule
\multicolumn{3}{c}{\textbf{COCO \ \ \ \ \ \ }} & \multicolumn{3}{c}{\textbf{\ \ \ \ Objects365}} \\
\midrule
airplane & $\rightarrow$ & plane & sneakers & $\rightarrow$ & running shoe \\
fire hydrant & $\rightarrow$ & hydrant & other shoes & $\rightarrow$ & shoe \\
suitcase & $\rightarrow$ & luggage & desk & $\rightarrow$ & table \\
skis & $\rightarrow$ & ski & street lights & $\rightarrow$ & street light \\
potted plant & $\rightarrow$ & plant & cabinet/shelf & $\rightarrow$ & cabinet \\
dining table & $\rightarrow$ & table & handbag/satchel & $\rightarrow$ & handbag \\
toilet & $\rightarrow$ & toilet bowl & picture/frame & $\rightarrow$ & picture \\
tv & $\rightarrow$ & television & gloves & $\rightarrow$ & glove \\
cell phone & $\rightarrow$ & smartphone & leather shoes & $\rightarrow$ & leather shoe \\
refrigerator & $\rightarrow$ & fridge & potted plant & $\rightarrow$ & plant \\
teddy bear & $\rightarrow$ & teddy & bowl/basin & $\rightarrow$ & bowl \\
& & & boots & $\rightarrow$ & boot \\
& & & monitor/tv & $\rightarrow$ & monitor \\
& & & trash bin can & $\rightarrow$ & can \\
& & & slippers & $\rightarrow$ & slipper \\
& & & barrel/bucket & $\rightarrow$ & barrel \\
& & & sandals & $\rightarrow$ & sandal \\
& & & pen/pencil & $\rightarrow$ & pencil \\
& & & wild bird & $\rightarrow$ & bird \\
& & & high heels & $\rightarrow$ & high heel \\
& & & cell phone & $\rightarrow$ & smartphone \\
& & & canned & $\rightarrow$ & drink \\
& & & lifesaver & $\rightarrow$ & life jacket \\
& & & awning & $\rightarrow$ & canopy \\
\bottomrule
\end{tabular}
\end{table}

To ensure compatibility with the label space of the RAM++ model, the categories from the COCO 2014 and O365 datasets must be appropriately mapped. The specific category correspondences with naming differences are summarized in Table \ref{tb:diff}. Categories not mentioned in the table remain unchanged, as their names are consistent with those in RAM++’s label space. As shown in the table, the mapping discrepancies can be categorized into three types: (1) singular vs. plural forms (\textit{e.g.}, ``gloves" $\rightarrow$ ``glove"); (2) synonymous expressions (\textit{e.g.}, ``refrigerator" $\rightarrow$ ``fridge"); and (3) near-synonyms expressions(\textit{e.g.}, ``desk" $\rightarrow$ ``table"). This structured mapping ensures that the annotated labels are consistent with RAM++’s predefined categories, thereby facilitating smoother training and evaluation processes. RAM++ produces per-class probability scores. Consistent with the original papers, we obtain the final predictions by applying suggested thresholds. The class-specific thresholds are available in \footnote{\url{https://github.com/xinyu1205/recognize-anything/blob/main/ram/data/ram_tag_list_threshold.txt}.}.

% \paragraph{Implementation}

% \paragraph{Category Mapping for RAM} 

\paragraph{Human Annotation} The human annotations referred to in this paper are the original annotations provided in the respective datasets. Below, we briefly introduce the workflows and methodologies related to image category annotation in each dataset.
\begin{itemize}
\item \textbf{MS-COCO} Dataset: With 91 object categories (the 2014 release was limited to 80) and a large number of images, it would be prohibitively expensive to have workers answer all these binary classification questions per image. Therefore, the authors adopted a hierarchical annotation strategy. The object categories were grouped into 11 super-categories. Annotators were first asked to determine whether any instance from a given super-category (\textit{e.g.}, ``animal") was present in the image. If present, they then annotated specific subordinate categories (\textit{e.g.}, ``dog", ``cat") within that super-category. To improve efficiency and accuracy, annotators placed category icons onto corresponding object instances in the image via a drag-and-drop interface. To ensure high recall of category labels, multiple annotators independently labeled each image, and the final determination of category presence was based on the union of all annotators’ responses. Each image was independently annotated by 8 annotators via Amazon Mechanical Turk. To ensure annotation accuracy, false positives identified during this stage were handled in the subsequent instance segmentation stage, where workers could indicate that no instance of a given category was present in the image.
\item \textbf{Objects365} Dataset: The annotation pipeline was designed to efficiently handle a large number of categories. The final dataset contains 365 object categories, selected based on frequency from an initial candidate set of 442 categories, and organized under 11 predefined super-categories (\textit{e.g.}, ``clothes", ``kitchen"). The annotation process began with a binary classification step to filter out iconic images or those containing none of the target objects. Images that passed this filter proceeded to the image-level tagging stage, where annotators identified one or more super-categories present in the image. This phased and super-category-driven approach significantly reduced the cognitive load on individual annotators, who only needed to be familiar with approximately 30-40 categories within a specific super-category, thereby ensuring feasibility and consistency in large-scale category annotation. The entire process was carried out by a professional team consisting of Annotators (who were required to complete a training course and pass an examination before starting annotation), Inspectors, and Examiners, working in coordination to ensure high-quality annotation results.
\end{itemize}
Overall, both COCO and Objects365 employed similar hierarchical annotation strategies, grouping fine-grained categories into super-categories and performing annotation in a staged manner. Furthermore, both datasets implemented validation mechanisms to ensure annotation accuracy.

%%%%%%%%%%%%%%%%%%%%%%%%%%%%%%%%%%%%%%%%%%%%%%%%%%%%%%%%%%%%%%%%%%%%%%%%%%%%%%%
%%%%%%%%%%%%%%%%%%%%%%%%%%%%%%%%%%%%%%%%%%%%%%%%%%%%%%%%%%%%%%%%%%%%%%%%%%%%%%%

\end{document}